\documentclass[letterpaper, 10 pt, conference]{ieeeconf}
\IEEEoverridecommandlockouts
\usepackage[nocompress,nospace]{cite}
\usepackage{amsmath,amssymb,amsfonts}
\usepackage{algorithmic}
\usepackage{graphicx}
\usepackage{textcomp}
\usepackage{xcolor}
\usepackage{placeins}
\usepackage{float}
\usepackage[hidelinks]{hyperref}

\usepackage{amsmath,amssymb,stmaryrd,mathtools}
\usepackage{xcolor}
\usepackage{xspace}
\usepackage{bbm}
\usepackage{graphicx}
\usepackage{subfig}
\usepackage{tcolorbox}
\usepackage{colortbl}
\usepackage{multirow}
\usepackage{booktabs}
\usepackage{wrapfig}
\usepackage{cleveref}

\crefname{section}{Sec.}{Secs.}
\Crefname{section}{Sec.}{Secs.}
\crefname{figure}{Fig.}{Figs.}
\Crefname{figure}{Fig.}{Figs.}
\crefname{table}{Tab.}{Tabs.}
\Crefname{table}{Tab.}{Tabs.}

\definecolor{dark_red}{RGB}{122, 0, 0}
\definecolor{coral}{RGB}{255, 119, 94}
\definecolor{pink_orange}{RGB}{255, 72, 126}
\definecolor{vibrant_pink}{RGB}{255, 0, 104}
\definecolor{pink_pink}{RGB}{255, 37, 153}
\definecolor{wine}{RGB}{204, 0, 102}

\definecolor{light_orange}{RGB}{255, 198, 107}
\definecolor{orange(sae/ece)}{rgb}{1.0, 0.49, 0.0}
\definecolor{dark_orange}{RGB}{216,92,0}

\definecolor{org-purp-0}{RGB}{165, 76, 0}
\definecolor{org-purp-1}{RGB}{250, 130, 28}
\definecolor{org-purp-2}{RGB}{226, 89, 68}
\definecolor{org-purp-3}{RGB}{206, 92, 124}
\definecolor{org-purp-4}{RGB}{116, 80, 146}
\definecolor{org-purp-5}{RGB}{110, 78, 157}

\definecolor{teal(sae/ece)}{rgb}{0, 0.47, 0.52}
\definecolor{aqua}{RGB}{52,172,139}
\definecolor{dark_aqua}{RGB}{35,115,93}
\definecolor{dark_green}{RGB}{0, 92, 34}

\definecolor{grape}{RGB}{112,48,160}
\definecolor{purple}{rgb}{0.74, 0.65, 1.0}
\definecolor{dark_purple}{rgb}{0.58, 0.0, 0.82}
\definecolor{periwinkle}{RGB}{191, 140, 230}

\definecolor{light_gray}{rgb}{0.9, 0.9, 0.9}
\definecolor{medium_gray}{rgb}{0.6, 0.6, 0.6} 
\definecolor{dark_gray}{rgb}{0.2, 0.2, 0.2} 

\definecolor{sky_blue}{RGB}{37, 166, 213}
\definecolor{light_blue}{rgb}{0.33, 0.80, 1}
\definecolor{dark_blue}{rgb}{0.098, 0.239, 0.52}
\definecolor{ocean}{RGB}{13, 121, 202}
\definecolor{light_ocean}{RGB}{18, 178, 235}
\definecolor{dark_ocean}{RGB}{10, 89, 148}
\definecolor{vibrant_blue}{RGB}{14, 120, 255}

\definecolor{dark_brown}{rgb}{0.3255, 0.004, 0.001}

\newcounter{qnum}
\newcounter{tnum}
\newcommand{\ourtitle}{RopeFormer\xspace}

\newcommand{\taska}{\texttt{Rope\_Swing}\xspace}
\newcommand{\taskb}{\texttt{Rope\_Twirl}\xspace}
\newcommand{\taskc}{\texttt{Rope\_Whip}\xspace}

\newif\ifshowchanges
\showchangestrue

\hypersetup{colorlinks=true,urlcolor=dark_orange,linkcolor=black,citecolor=black}
\showchangesfalse
\def\BibTeX{{\rm B\kern-.05em{\sc i\kern-.025em b}\kern-.08em
    T\kern-.1667em\lower.7ex\hbox{E}\kern-.125emX}}

\begin{document}
\bstctlcite{BSTcontrol}

\title{\LARGE \bf
RopeFormer: Cross-Trial Adaptation from Interaction History for Dynamic Rope Manipulation
}

\author{Menglin Wu$^{*,1,2}$, Kaixiang Yao$^{*,1,3}$, Shangbo Luan$^{1,4}$, Masayoshi Tomizuka$^{1}$, and Yuxin Chen$^{1}$\\
{\small $^{1}$\textit{University of California, Berkeley}\quad
$^{2}$\textit{Xi'an Jiaotong University}}\\
{\small $^{3}$\textit{Southern University of Science and Technology}\quad
$^{4}$\textit{Peking University}}\\
{\small $^{*}$Equal contribution.}
}

\maketitle
\thispagestyle{empty}
\pagestyle{empty}

\global\csname @topnum\endcsname 0
\global\csname @botnum\endcsname 0

\begin{abstract}
Dynamic rope manipulation is highly sensitive to unknown object dynamics:
the same robot motion can produce substantially different responses across
ropes, while explicitly identifying the relevant physical properties is
difficult. We present RopeFormer, a history-conditioned framework that uses
prior task interaction as context for subsequent control. The policy retains
cross-trial action--response history while keeping its weights fixed and
requires no explicit online rope-parameter estimation. In matched simulation
evaluations across sustained single-arm rotation, bimanual rotation, and
transient whipping, retaining context improves subsequent control relative to
resetting the same checkpoint, with the benefit varying across rope dynamics
and observation settings. We further deploy the frozen policies on a Unitree
H1-2 with previously unseen physical ropes. From T1 to T3, target-acquisition
time decreases by 30.9\% for \taska and 33.9\% for \taskb, while mean
\taskc target hits increase from 0.2 to 2.3 out of three. These results show that prior interaction can provide effective control context for dynamic deformable-object manipulation. Robot videos, code, and data are available at \url{https://ropeformer.github.io/}.
\end{abstract}


\begin{figure}[t]
    \centering
    \includegraphics[width=\columnwidth]{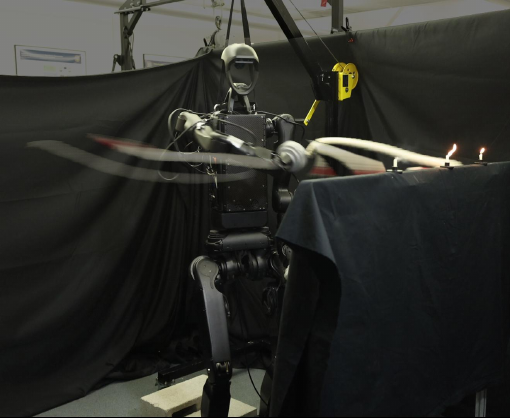}
    \caption{Each dynamic interaction is both a task execution and a
    source of information about the rope. RopeFormer retains
    action--response history across repeated trials and uses it to
    condition subsequent control with fixed policy weights. We study
    this principle across three dynamic rope-manipulation tasks; shown
    here is a representative \taskc deployment on the Unitree H1-2
    against physical candle targets.}
    \label{fig:teaser}
\end{figure}

\section{Introduction}
\label{sec:introduction}
Dynamic rope manipulation enables robots to perform fast and dexterous behaviors that exploit, rather than avoid, the dynamics of deformable objects. For example, \Cref{fig:teaser} shows a humanoid whipping a rope to extinguish physical candle targets. Such dynamic behaviors are particularly challenging because ropes are soft, highly deformable, and underactuated. Their motion depends on complex interactions among mass distribution, bending stiffness, damping, geometry, and configuration, and different ropes can therefore respond substantially differently to the same robot motion~\cite{lostarc2020,irp2022,freeend2024}.
This sensitivity is especially pronounced during fast manipulation, where small differences in rope dynamics can lead to large differences in motion and task outcome.

Existing approaches often address this uncertainty by explicitly modeling or identifying rope dynamics. Physical interactions can be used to calibrate simulation or estimate rope properties before control~\cite{real2sim2real2022,genorm2023,wiggle2026}, while repeated-task methods construct local action--response models or explicit action updates from previous executions~\cite{irp2022,ilc2026}. However, accurately modeling rope dynamics is itself difficult: relevant physical properties can be hard to identify, simplified models cannot capture all real-world effects, and modeling errors can lead to a substantial sim-to-real gap. At the same time, interaction with the rope naturally reveals how it responds to robot actions, providing information about the underlying task dynamics without requiring them to be explicitly parameterized. \emph{Instead of explicitly modeling rope dynamics that are difficult to identify and transfer, can a robot learn task-relevant dynamics implicitly from its interaction history and use them to improve subsequent control?}

We introduce \textbf{RopeFormer}, a framework for cross-trial history-conditioned adaptation in dynamic rope manipulation. Rather than explicitly estimating physical parameters, RopeFormer directly conditions control on previous action--response experience. A Transformer-XL policy retains the ordered history of robot states, actions, task information, and observed rope motion across repeated trials. During training, multiple attempts share randomized rope--actuator dynamics: the physical state is reset between trials while policy context is retained, allowing experience from earlier attempts to inform later control. At deployment, the policy receives no explicit rope parameters and its weights remain fixed; adaptation occurs through retained context.

We evaluate RopeFormer across three qualitatively distinct dynamic rope-manipulation tasks: single-arm sustained rotation (\taska), bimanual sustained rotation (\taskb), and transient target-line whipping (\taskc). Through controlled simulation experiments, we isolate the effect of cross-trial interaction history and study how its benefit varies across rope dynamics and observation settings.
We further deploy the learned policies on a Unitree H1-2 humanoid with previously unseen physical ropes. Across all three tasks, the robot improves over repeated interactions while keeping its policy fixed, demonstrating that prior action--response experience can provide effective context for dynamic deformable-object control.

Our contributions are threefold:
\begin{itemize}
    \item We formulate repeated dynamic manipulation as a history-conditioned control problem in which previous action--response experience provides task-relevant information for subsequent control.

    \item We introduce a multi-trial training framework and Transformer-XL policy that retain ordered interaction context across physical trial resets, enabling adaptation through experience with fixed policy parameters.

    \item We demonstrate the approach on sustained single-arm rotation, coordinated bimanual rotation, and transient whipping through controlled simulation studies and repeated real-world experiments on a Unitree H1-2.
\end{itemize}

\section{Related Work}

\subsection{Dynamic DLO Manipulation under Unknown Dynamics}

Dynamic manipulation of deformable linear objects (DLOs) has been studied
through
self-supervised interaction, simulation-based policy learning, learned dynamics,
and model-based control \cite{dloSurvey2026}. \emph{Robots of the Lost Arc}
learns to manipulate fixed-endpoint cables from self-supervised physical
interaction, while subsequent work extends dynamic manipulation to free-end
cables \cite{lostarc2020,freeend2024}. Simulation-based learning has also been
explored through physics-based dynamic DLO simulation
\cite{chenDynamicDLO2023} and model-free goal-conditioned dexterous control
\cite{dexdlo2024}. A recurring challenge is variation across ropes and
rope--environment conditions: changes in rope geometry and material properties
can substantially alter the outcome of the same action.

Several methods use physical interaction to infer this variation before or
during control. Real2Sim2Real calibrates simulation from physical cable
trajectories before generating training data for planar casting
\cite{real2sim2real2022}, while GenORM combines a parameter-aware policy with
estimates of the physical properties of a new rope \cite{genorm2023}.
\emph{Wiggle and Go!} uses a brief diagnostic motion to identify descriptive
rope parameters that subsequently inform goal-conditioned dynamic manipulation
\cite{wiggle2026}. Dynamics-aware approaches also combine physical structure
with learned control, including physics-informed self-supervised learning
\cite{spid2026} and physics-informed test-time adaptation for 3-D rope
manipulation \cite{dynamic3d2025}. Other approaches exploit repeated task
execution directly. Iterative Residual Policy predicts how action perturbations
change a previously observed deformable-object trajectory and searches for an
improved action on the next attempt \cite{irp2022}, while Task-Level Iterative
Learning Control constructs a local inverse model that maps task-space error to
subsequent command updates \cite{ilc2026}. These methods therefore use
interaction to obtain explicit model or parameter estimates, or to compute
explicit action updates. In contrast, \ourtitle retains the ordered
state--action--response history directly as policy context, learning the mapping
from prior interaction to subsequent control jointly with the task policy.

\subsection{History-Conditioned Adaptation}

Interaction history can also provide context for latent task and system
properties. RL$^2$ showed that a recurrent policy can retain information across
episodes of the same unseen task and implement a learned adaptation procedure
through its hidden state \cite{rl22016}. ICRT conditions robot control on
sensorimotor trajectories provided as in-context task demonstrations
\cite{icrt2025}. In robotics, UP-OSI uses recent state--action history to
estimate dynamics parameters for a universal policy \cite{unknown2017}, while
Rapid Motor Adaptation and related manipulator work learn task-relevant latent
dynamics from recent interaction \cite{rma2021,rmaarms2023}. HORA applies rapid
adaptation to in-hand object rotation, using proprioceptive history to adjust
control to objects with different physical properties \cite{hora2023}.
Interaction history can also support system identification; \emph{Dynamics as
Prompts} uses previous trajectories as context for adapting simulated dynamics
toward a target system \cite{dynamicsPrompts2024}, while In-Context World
Modeling infers system variables from short self-generated interactions without
parameter updates \cite{worldmodel2026}. Together, these works establish
interaction history as a useful source of adaptation. \ourtitle instead
conditions control directly on ordered interaction history for an external,
highly underactuated deformable object.

Gated Memory Policy explicitly studies in-trial and cross-trial memory in
manipulation, learning when and what history to recall across repeated attempts
\cite{gmp2026}. Most closely related to our policy design, LocoFormer studies
long-context adaptation for locomotion by extending Transformer-XL context
across trial boundaries, allowing earlier interactions to influence later
control under previously unseen robot morphologies and dynamics
\cite{locoformer2025}. \ourtitle adopts this cross-trial long-context principle
for dynamic rope manipulation, where interaction history couples robot motion
with the delayed and spatially distributed response of an external deformable
object. We study whether this state--action--response history can directly serve
as control context over repeated rope-manipulation trials.


\section{Method}
\label{sec:method}

\begin{figure*}[t]
    \centering
    \includegraphics[width=\textwidth]{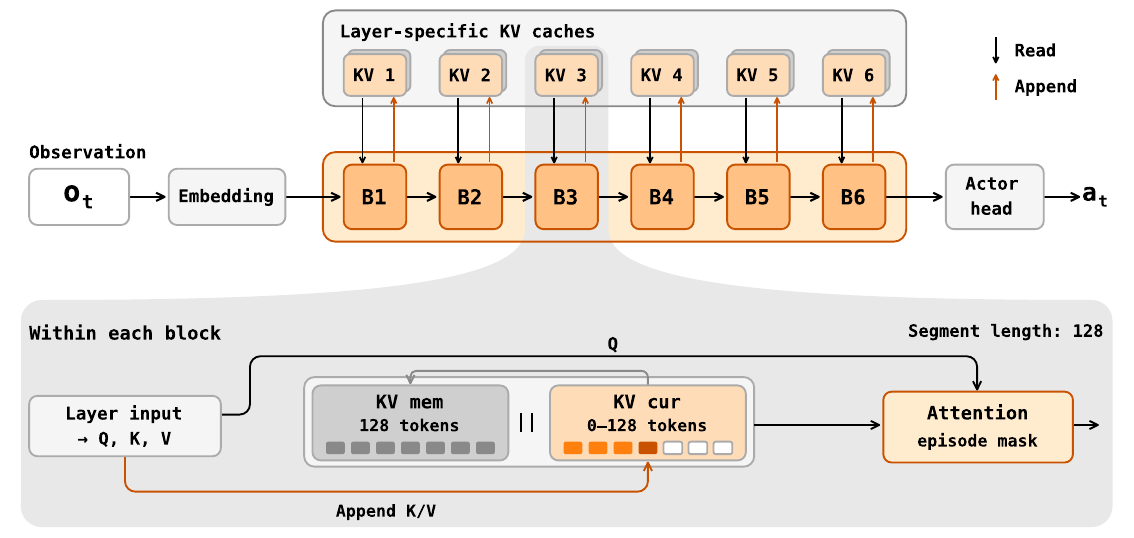}
    \caption{RopeFormer controls ropes with unknown dynamics by conditioning each action on the history of robot actions and observed rope responses. At step $t$, the observation $o_t$ contains robot states, measured rope-point positions, the task command or target, and the previous action. An embedding, six Transformer-XL blocks, and an actor head map this input and the retained context to the joint-target command $a_t$. Context persists across physical trial resets within an episode, allowing later attempts to use earlier interactions while policy weights remain fixed. The expanded block shows streaming inference shared by all six layers. The current query $Q$ attends to concatenated KV mem (preceding segment) and KV cur (current segment), with masking to exclude other episodes. New keys and values enter KV cur before attention. After all blocks finish the step, a full 128-token KV cur replaces KV mem and is cleared. KV mem is initially empty; segment rollover is distinct from a trial reset.}
    \label{fig:pipeline}
\end{figure*}

RopeFormer addresses uncertainty in rope dynamics by using past robot actions
and rope responses as context for the next control decision. It does not
explicitly estimate physical parameters online. Instead, a history-conditioned
policy learns from repeated interactions in randomized simulation and retains
its interaction context while keeping its weights fixed during execution (\Cref{fig:pipeline}).
We first define the task and repeated-interaction setting (\Cref{sec:method-problem}),
then the history-conditioned policy and streaming implementation  (\Cref{sec:method-policy}), and finally the multi-trial training protocol (\Cref{sec:method-training}).

\subsection{Problem Formulation}
\label{sec:method-problem}

\begin{figure*}[t]
    \centering
    \includegraphics[width=\textwidth]{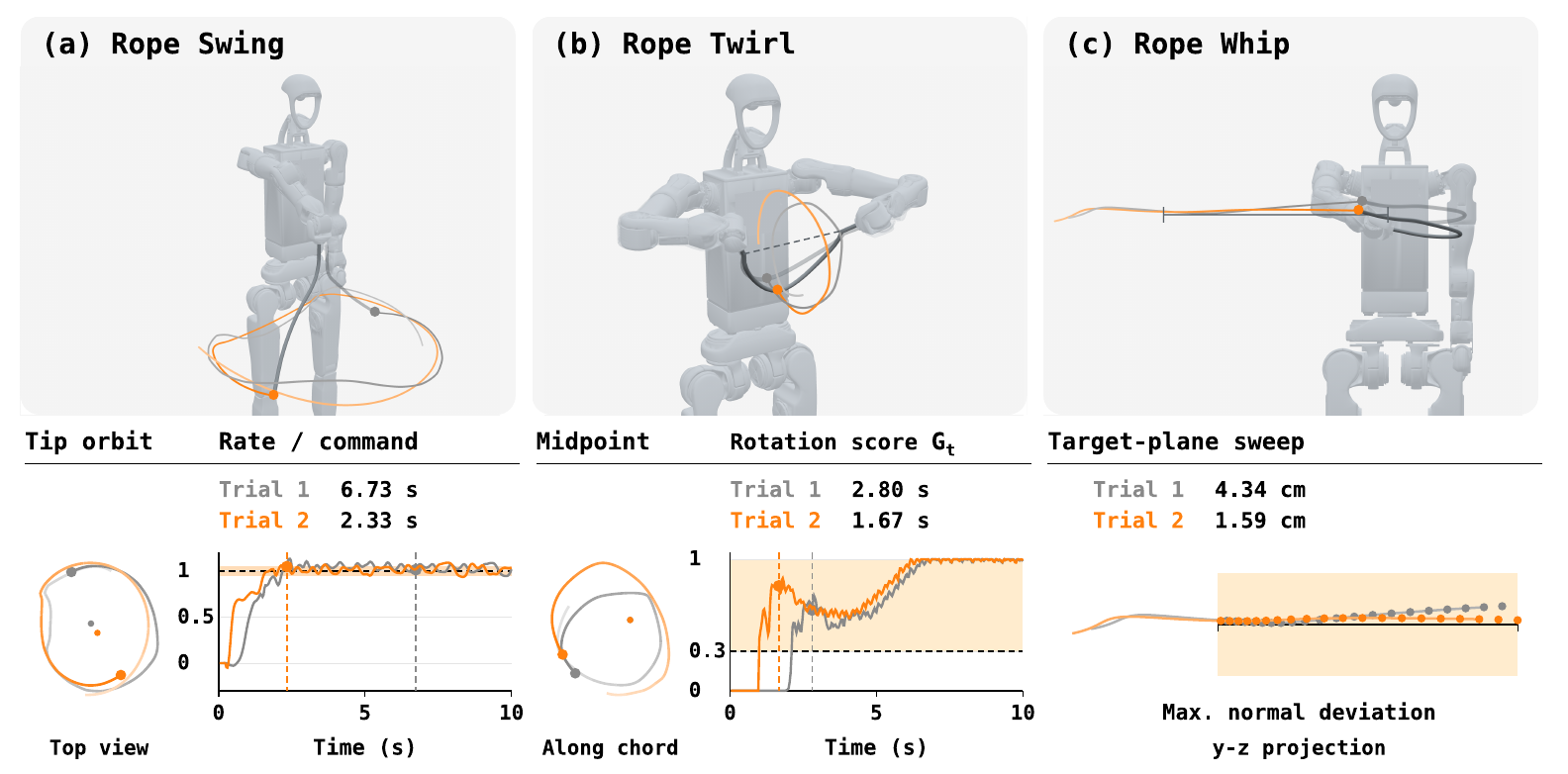}
    \caption{The three manipulation tasks and selected paired-trial examples in simulation. Swing rotates the free tip, Twirl rotates the midpoint with both ends held, and Whip sweeps the tip along a target line. Grey and orange denote trials 1 and 2, respectively. Translucent trial-1 poses are overlaid on trial 2 at the same elapsed time. Swing and Twirl poses are shown at trial-2 acquisition, with tip or midpoint orbits in the insets. Curve markers indicate the end of each trial's first qualifying acquisition window. Whip shows the target-line sweep in the $y$--$z$ scoring plane, with counted samples and maximum normal deviations below. Metrics: Sec.~\ref{sec:evaluation-setup}; aggregate comparisons: Sec.~\ref{sec:q1-cross-trial-benefit}.}
    \label{fig:task-overview}
\end{figure*}

We consider three dynamic rope-manipulation tasks, illustrated in~\Cref{fig:task-overview}.
In \taska, the right hand holds one rope end and drives the free tip
into sustained rotation at a commanded signed angular velocity.
In \taskb, the robot holds both rope ends and coordinates both arms
to rotate the rope midpoint about the axis defined by the line joining the
hands at a commanded angular velocity.
In \taskc, the right hand holds one end and drives the free tip along
a prescribed line segment in a task plane. The first two tasks require
sustained rotational motion, whereas \taskc requires a transient sweep with
both target coverage and trajectory accuracy.

At control step $t$, the policy receives an observation $o_t$ containing
normalized robot joint states, measured rope-point positions, the task command
or target, and the previous action. Command phase and additional controller
state are included where required by the task. The output $a_t$ specifies
a joint-target command for the controlled arm or arms. The actor observes
neither the full rope state nor its physical parameters. Rope-point positions
observed at a single time step do not fully specify the rope's motion or the
dynamics governing its response to robot actions.
The control problem is to use the available observations and interaction
history to produce the desired rope motion under unknown dynamics.

We study this problem over repeated physical trials. A \emph{trial} is one
attempt at the task, from initialization until a termination or time limit.
An \emph{episode} groups successive trials under a sampled rope--actuator
condition. Resetting the robot--rope state between trials does not, by itself,
remove information about that condition. We therefore retain policy context
across trial resets and isolate it when a new episode begins. This gives
later actions access to earlier task interactions without online weight
updates. The three tasks share this formulation, network architecture, and
training protocol, but use separately trained policies.

\subsection{History-Conditioned Policy}
\label{sec:method-policy}

\paragraph{From interaction history to actions}
To connect commanded motion with the rope's observed response, the policy
processes observations in temporal order. A feed-forward encoder maps the
observation at each control step to a single token. A Transformer-XL (TXL) backbone
combines this token with cached representations of earlier observations to
produce a hidden representation $h_t$. An actor head maps $h_t$ to a diagonal
Gaussian action distribution. At execution, its mean is clipped and mapped
to joint targets. The resulting rope motion enters subsequent observations,
closing the feedback loop. Thus, history conditions the action directly,
without an intermediate physical-parameter estimate.

\paragraph{Retaining context across segments and trials}
We adopt Transformer-XL segment recurrence~\cite{transformerxl2019}, following
its use for history-conditioned control in LocoFormer~\cite{locoformer2025}.
Let $z$ index segments of $L$ control frames, with $H_z^0$ collecting their
encoded tokens. At layer $\ell$, the current layer inputs $H_z^{\ell-1}$
attend to both the current segment and cached inputs $H_{z-1}^{\ell-1}$ from
the preceding segment. With concatenation denoted by $[\,;\,]$ and
stop-gradient by $\operatorname{SG}$, the recurrence is
\begin{equation}
\begin{aligned}
\widetilde H_z^{\ell-1}
  &= [\operatorname{SG}(H_{z-1}^{\ell-1});H_z^{\ell-1}],\\
H_z^\ell
  &= \operatorname{Block}_{\ell}
     (H_z^{\ell-1},\widetilde H_z^{\ell-1};\mathcal M_z).
\end{aligned}
\label{eq:segment-recurrence}
\end{equation}
Here, $\operatorname{Block}_{\ell}$ includes normalization, position-aware
attention, feed-forward operations, and residual connections. Queries come
from the current segment, while keys and values come from the concatenated
representations. The mask $\mathcal M_z$ enforces causality and excludes
tokens from other episodes. Cached representations are detached during
training but remain available as control context.

Deeper-layer representations can carry information from earlier segments,
extending the temporal dependency beyond one segment. Crucially, a physical
trial reset does not interrupt this recurrence within an episode. Segment
boundaries organize computation, whereas trial boundaries restart physical
motion. Keeping these boundaries separate allows prior action--response
information to condition a new attempt. The retained context is finite:
whether it spans part of a trial or several trials depends on their duration.

\paragraph{Streaming execution}
The six-layer, eight-head backbone uses segments of $L=128$ frames. This
sets the per-layer segment length, not a 128-frame limit on the entire
network's temporal dependency. For streaming execution, the layer-input
memory in Eq.~\ref{eq:segment-recurrence} is represented by cached projected
keys and values (KV), as shown in~\Cref{fig:pipeline}. Each layer
processes one new token using its previous-segment cache and the available
part of the current segment. Completed segments replace the preceding cache,
so memory remains bounded and an action is produced at every control step.
Reusing KV avoids recomputing the full history, although a longer cache still
increases attention work. Execution updates the context with fixed policy weights.

\subsection{Multi-Trial Policy Training}
\label{sec:method-training}

\paragraph{Learning from repeated responses}
The training protocol gives the policy opportunities to use the memory
described above. Within each episode, the robot makes multiple attempts,
resetting the physical state between trials while normally retaining the
sampled dynamics and policy context. A new episode samples a new condition
and excludes the preceding episode's context. Across episodes, the policy
therefore encounters different action--response relationships; within an
episode, earlier responses remain available when selecting subsequent actions.
This protocol supplies context for repeated control, rather than assuming
that every successive trial must improve.

We train in Newton~\cite{newton2025} with randomized rope and actuator
dynamics.
Rope variations include mass density, bending stiffness, damping, and drag,
with task-specific variation in length or added mass. Actuator variations
include tracking response, command delay, and motion limits. These variations
cover both rope response and differences between commanded and realized arm
motion. Episode-level bias and frame-level noise perturb rope observations,
with temporary point dropout in Swing and Whip. Task-specific curricula vary
dynamics or initialization and target difficulty. Some curricula also switch
rope properties within a trial without clearing context; such switches are
not a requirement of the shared multi-trial formulation. Twirl initializes the actor by behavioral cloning before reinforcement learning.

\paragraph{Task rewards and optimization}
Rewards express the motion objectives in~\Cref{sec:method-problem}.
For Swing, angular-rate tracking is combined with planarity, rope extension,
and orbit placement. Twirl rewards rotation rate, radius, and direction,
together with a level hand-to-hand line and body clearance. Whip rewards new
target coverage weighted by proximity, together with completion and settling.
Task-specific shaping and penalties discourage undesired arm and rope motion.
These simulator-computed rewards guide training without exposing the
simulator-only information to the actor. Evaluation criteria are specified
separately in~\Cref{sec:evaluation-setup}.

We optimize the recurrent actor--critic with proximal policy optimization
(PPO)~\cite{ppo2017}, using ordered segments and their cached starting context.
An asymmetric critic additionally receives simulator-only rope and actuator
parameters, while the actor uses only its observation history. Generalized
advantage estimation and value bootstrapping stop at trial resets even when
policy context is retained. Actions in each trial are therefore optimized
for that trial's task return, not for rewards obtained in later trials.
The resulting interaction history remains available to condition later
control. The policy can thus use experience from earlier attempts without
explicitly optimizing those earlier actions for their information value
in subsequent trials.

\section{Experiments}
\label{sec:experiments}

\begingroup
\raggedbottom
\clubpenalty=10000
\widowpenalty=10000

We organize the evaluation around three questions: whether retained
cross-trial history improves control (Q1), how that benefit varies with
dynamics and observation settings (Q2), and whether repeated improvement
persists on hardware (Q3). Q1--Q2 use matched simulation comparisons,
while Q3 evaluates previously unseen physical ropes.

\begin{figure}[!t]
    \centering
    \includegraphics[width=\columnwidth]{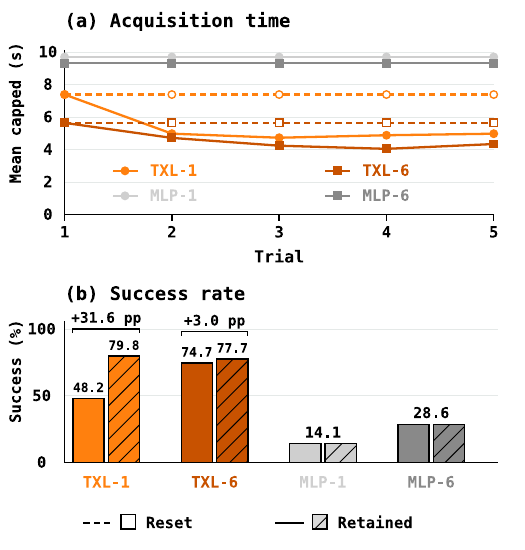}
    \caption{Simulated single-arm rotation (\taska), 384 ropes and five trials. TXL/MLP denote Transformer-XL/multilayer perceptron; 1/6 denotes observed rope points. Matched checkpoints retain or reset context across trials. (a) Mean target-acquisition time (failures: 10~s): solid/filled is Retained, dashed/open is Reset. (b) Trial-2--5 mean success: plain bars are Reset, hatched bars Retained; annotations give percentage-point gains. MLP conditions overlap. Retaining context yields faster acquisition and higher success for both TXLs, especially TXL-1.}
    \label{fig:swing-results}
    \label{fig:sim-results}
\end{figure}
\subsection{Evaluation Setup}
\label{sec:evaluation-setup}

\taska uses one arm for sustained rotation, \taskb uses two arms to rotate a
rope about the hand-to-hand chord, and \taskc uses one arm for a transient
sweep toward a target path. The tasks use separately trained policies with
weights frozen throughout evaluation. \Cref{fig:task-overview}
illustrates the three tasks and representative task metrics.

For the two rotation tasks, we compare TXL and multilayer perceptron (MLP)
policies observing one or six rope keypoints, denoted TXL-1, TXL-6, MLP-1,
and MLP-6. The single point is
the free tip for \taska and the midpoint for \taskb. Our primary control
compares the \emph{same} TXL checkpoint with context either retained across
trial boundaries or reset at every boundary; both conditions retain
within-trial history. The MLP uses an eight-frame observation stack that is
reset at every trial, so it provides a non-recurrent reference with no access
to previous trials.

\paragraph{Simulation protocol and metrics}
For each task, we generate a fixed evaluation set of 384 ropes using
task-specific parameter samplers. Linear density is drawn uniformly, while
bending stiffness, damping, and air-drag coefficients are drawn log-uniformly;
\taska and \taskc additionally randomize added masses. \taska samples are
screened for a feasible rotation-command range. Free rope length spans
0.35--0.62~m over nine settings for \taska and 0.50--0.65~m over seven
settings for \taskb, and is fixed at 0.90~m for \taskc. Bending-stiffness
ranges are 1--30, 0.75--18, and 0.5--100~N$\cdot$m/rad, respectively.
Rope properties remain fixed within each five-trial sequence, and all
comparisons within a task reuse the same evaluation set.

Each rotation task uses five 10-s trials per rope at 30~Hz. Retained/reset pairs share the checkpoint, rope and target manifest,
initial states, and observation-noise settings. We evaluate one training seed per configuration.
We estimate the start of steady-state execution with a 1-s
sliding window: acquisition time is the end of the first window in which at
least 70\% of frames satisfy the task-specific criteria. Meeting this criterion defines success; otherwise acquisition time is capped at 10~s, with later-trial comparisons averaged over T2--T5.

For \taska, qualifying frames require absolute relative error in signed angular velocity below 5\%, axis
tilt at most $15^\circ$, and hand displacement below 0.15~m. The orbit center
must lie below the hand by at least half the reference steady-cone droop, with
a 0.05-m minimum. For \taskb, qualification uses
\begin{equation}
G_t=g_{\omega,t}g_{r,t}g_{\kappa,t}g_{\theta,t}g_{c,t}>0.30,
\label{eq:twirl-eval-score}
\end{equation}
which combines rate $g_{\omega,t}$, effective radius $g_{r,t}$, directionality $g_{\kappa,t}$, chord tilt $g_{\theta,t}$, and body
clearance $g_{c,t}$ over a 30-frame motion window.
For relative angular-rate error
$e_t$, $g_{\omega,t}=\exp[-(e_t/0.15)^2/2]$. The remaining gates are clipped
linear ramps: effective radius normalized by free arc length from 0.1333 to
0.30, signed directionality from 0.50 to 0.90, maximum chord tilt from
$15^\circ$ down to $5^\circ$, and minimum clearance from 0 to 0.02~m.
Midpoint phase is measured in a chord-relative frame; directionality is signed
net phase advance divided by total absolute phase advance, and effective radius
is derived from swept area per phase advance. For \taskc, we evaluate the final TXL-1 checkpoint at seven target inclinations from
$0^\circ$ to $90^\circ$ in $15^\circ$ increments, using target
lengths of 0.5--0.8~m, and five 2.5-s trials per angle. The metric $d_{\max}$
is the maximum normal deviation of the free-tip projection in the pelvis
$y$--$z$ plane over counted frames within a 12-cm corridor; valid 
segments require finite deviation and at least 85\% tangential coverage.

\subsection{Q1: Does Cross-Trial History Improve Control?}
\label{sec:q1-cross-trial-benefit}
\label{sec:simulation-results}

For \taska, averaged over T2--T5 (\Cref{fig:swing-results}), retaining context decreases TXL-1 mean capped acquisition time from 7.39 to 4.89~s and increases success from 48.2\% to 79.8\%. TXL-6 decreases acquisition time from 5.64 to 4.34~s, while its success rate rises from 74.7\% to 77.7\%. The corresponding MLP retained/reset conditions coincide at the displayed precision. Under the matched design above, the TXL differences isolate access to
prior-trial context.

\begin{figure}[t]
    \centering
    \includegraphics[width=\columnwidth]{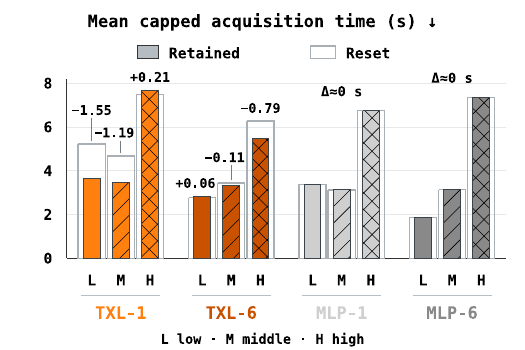}
    \caption{Simulated bimanual rotation (\taskb): trial-2--5 mean acquisition time (failures: 10~s; lower is better). TXL/MLP denote Transformer-XL/multilayer perceptron; 1/6 denotes observed rope points. Filled bars retain cross-trial context; grey outlines reset it for matched checkpoints and ropes. Plain/diagonal/crosshatched fills denote low/middle/high stiffness (128 ropes each). Annotations give $\Delta=\mathrm{Retained}-\mathrm{Reset}$ (s); negative values mean faster acquisition. TXL-1 benefits most at low/middle stiffness, TXL-6 at high stiffness; MLP conditions overlap.}
    \label{fig:twirl-results}
\end{figure}
For \taskb, averaged over all 384 ropes, retaining context decreases
T2--T5 acquisition time from 5.79 to 4.95~s for TXL-1 and
from 4.16 to 3.89~s for TXL-6 (\Cref{fig:twirl-results}). MLP results remain unchanged at displayed precision.

\begin{figure}[!tb]
    \centering
    \includegraphics[width=\columnwidth]{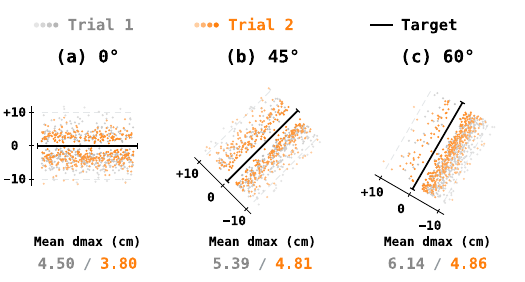}
    \caption{Simulated target-line sweeps (\taskc) with a fixed one-point Transformer-XL policy retaining context. Grey/orange show trials 1/2 for ropes scored in both trials ($n=377/375/369$ at $0/45/60^\circ$). Normal offsets encode maximum tip-path deviation $d_{\max}$ (cm), with sign indicating the side of deviation; along-line positions index ropes. Lighter points indicate larger deviations; numbers below panels are paired-set means of unsigned $d_{\max}$. Mean deviation decreases at all three displayed angles, most at $60^\circ$.}
    \label{fig:whip-results}
\end{figure}
\begin{figure*}[t]
    \centering
    \raisebox{-5pt}[\height][\depth]{%
        \includegraphics[width=\textwidth]{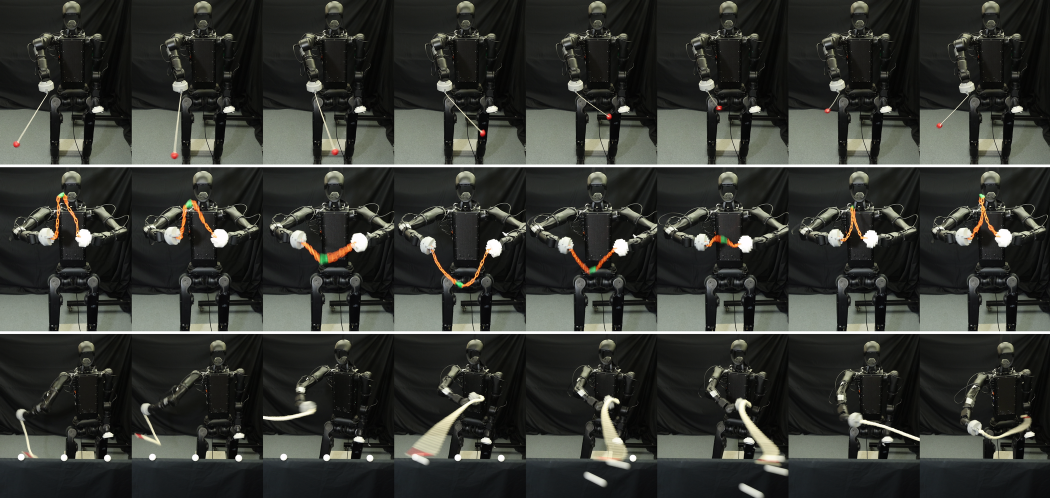}}
    \caption{Representative real-robot execution sequences for \taska, \taskb,
    and \taskc (top to bottom). Frames progress from left to right within each
    row.}
    \label{fig:hardware-executions}
\end{figure*}
For \taskc, comparisons on jointly valid retained/reset pairs yield mean
T2--T5 reductions in maximum normal deviation of 0.07, 0.41, and 0.53~cm
over the overlapping angle groups
$\{0,15,30\}^\circ$, $\{30,45,60\}^\circ$, and
$\{60,75,90\}^\circ$, respectively, while all-record target-length coverage
remains above 95\% in both conditions. Selected retained-context T1/T2 distributions are shown in~\Cref{fig:whip-results}.

Retained context improves aggregate performance, but unevenly across
conditions, motivating Q2.

\subsection{Q2: How Does History Value Depend on Dynamics and Observations?}
\label{sec:q2-context-dependence}

\Cref{fig:twirl-results} stratifies the 384 \taskb ropes into three
equally sized bending-stiffness groups (128 ropes each), with cut points at
approximately 2.2556 and 6.7272~N$\cdot$m/rad. The high-stiffness group has
the longest mean acquisition time for every architecture under both memory
conditions, although other rope properties also vary within each stratum.

With one observed rope point, TXL-1 acquires 1.55~s and 1.19~s sooner with
retained context in the low- and middle-stiffness groups, but is 0.21~s slower
in the high-stiffness group. With six observed points, TXL-6 changes little in
the low and middle groups and acquires 0.79~s sooner in the high-stiffness group. The MLP conditions are effectively unchanged by the retained/reset label. Absolute performance ordering also depends on architecture and stiffness: in the low-stiffness group, MLP-6 reaches 1.86~s versus 2.84~s for retained TXL-6, whereas in the high-stiffness group retained TXL-6 reaches 5.49~s
versus 7.36~s for MLP-6. Thus, the retained-context benefit varies across observation settings and
the evaluated dynamics regimes; the overall TXL gains persist when the same
trajectories are re-scored at $G_t>0.40$ and $G_t>0.50$.

\subsection{Q3: Does Repeated Improvement Persist on Hardware?}
\label{sec:real-robot-results}
\label{sec:real-robot-setup}

\paragraph{Deployment and repeated-trial protocol}
We deploy a frozen TXL-1 policy for each task on a Unitree H1-2
(\Cref{fig:hardware-executions}). Two ZED 2i cameras operate at 720p/60 Hz; a marker is detected in one RGB view from
each camera, and calibrated cross-camera rays are triangulated to recover its
3-D position. The marker is placed at the free tip for \taska and \taskc and at
the rope midpoint for \taskb. The policy loop runs at 30 Hz and the low-level H1 command publisher at 250 Hz. Measured rope properties are not policy inputs.

\begin{figure}[!tb]
    \centering
    \includegraphics[width=0.94\columnwidth]{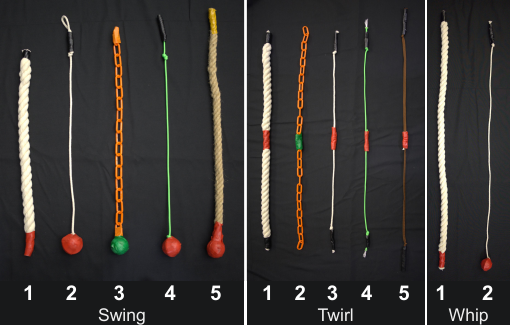}
    \caption{Physical ropes used in the real-robot experiments, from left to
    right: Rope Swing, Rope Twirl, and Rope Whip. Numbers identify rope
    instances within each task.}
    \label{fig:hardware-rope-suite}
\end{figure}

For \taska and \taskb, we evaluate five physical ropes (\Cref{fig:hardware-rope-suite}) with three independent three-trial blocks per rope. Each block begins with empty cross-trial context at T1 and retains context through T2 and T3, giving 45 trials per task. Between
trials, policy execution and memory updates pause while the robot--rope system
is returned to its prescribed initial condition and allowed to become
stationary. Reported T1--T3 aggregates first average the three blocks within
each rope and then average across the five ropes. The \taskc hardware study
uses two physical ropes and five target-height groups per rope, giving ten
three-trial groups and 30 task outcomes; target heights span 115--140~cm. Each group starts T1 with empty context; context accumulates online, is
retained through T2--T3, and is cleared between groups.

\paragraph{Sustained rotation: convergence and tracking quality}
For the two rotation tasks, we measure both how quickly the commanded motion is
acquired and how accurately it is sustained. The hardware estimator uses a
2-s rolling window. For \taska, angular velocity is estimated by rolling
regression of unwrapped tip phase about a slowly varying orbit center. For
\taskb, the midpoint is projected into the plane orthogonal to the
instantaneous hand-to-hand axis and a joint sinusoidal fit to the transverse
coordinates estimates frequency. Let $\hat{\omega}(t)$ and $\omega^*$ denote the estimated and commanded
angular velocities, and $\mathcal V$ the valid estimator times. The median
normalized error (MNE) is
\begin{equation}
\begin{aligned}
 e_\omega(t)&=\frac{|\hat\omega(t)-\omega^\ast|}{|\omega^\ast|},\\
 \mathrm{MNE}&=\underset{t\in\mathcal V}{\operatorname{median}}e_\omega(t)\times100\%.
\end{aligned}
\label{eq:real-mne}
\end{equation}
Target-acquisition time (TAT) is the earliest valid time with
$e_\omega(t)\le0.05$ for which at least 90\% of valid estimates in the
following 1~s also satisfy this bound. Values near 2~s correspond to
qualification as soon as the rolling estimator becomes available. These
hardware metrics are deployment-specific; the simulation acquisition metric
in \Cref{sec:evaluation-setup} additionally uses task-specific geometric
and clearance criteria.

\begin{figure}[!tb]
    \centering
    \includegraphics[width=\linewidth]{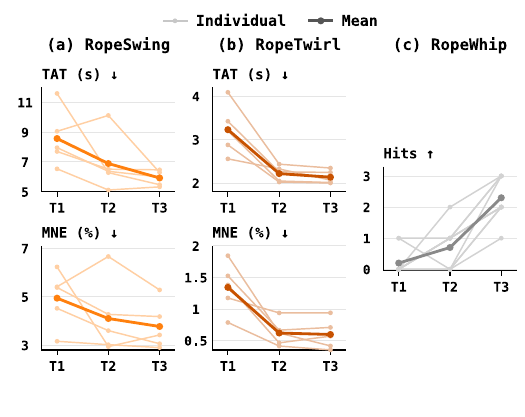}
    \caption{Physical-rope performance with fixed weights and context retained from T1 through T3. For \taska/\taskb, TAT is target-rotation acquisition time (s); MNE is within-trial median absolute relative angular-rate error (\%); both are lower-is-better. Thin traces average three sequences per rope; thick traces average five ropes. For \taskc, Hits counts balls struck out of three (higher-is-better); thin traces show ten rope--height groups across two ropes, thick traces their mean. T1 starts with empty context. Repetition reduces mean rotation acquisition time and error, and increases mean Whip hits.}
    \label{fig:hardware-results}
\end{figure}

\Cref{fig:hardware-results} shows consistent repeated improvement in the
two sustained-rotation tasks. For \taska, mean TAT decreases from 8.57 to 6.89
to 5.92~s over T1--T3, a 30.9\% reduction from T1 to T3, while MNE decreases
from 4.94\% to 4.10\% to 3.77\%, a 23.7\% reduction. For \taskb, with a
commanded frequency of 2.6~Hz, mean TAT decreases from 3.229 to 2.216 to
2.134~s, a 33.9\% reduction, while MNE decreases from 1.342\% to 0.619\% to
0.596\%, a 55.6\% reduction. Every evaluated rope has lower T3 than T1 TAT and
MNE in both rotation tasks. Most of the \taskb TAT improvement occurs between
T1 and T2, with later trials near the estimator floor.

\paragraph{Transient physical effect in \taskc}
\taskc tests whether repeated interaction also improves a transient
physical outcome. With three ping-pong balls placed along the target line,
mean hits across ten formal groups rise from 0.2 to 0.7 to 2.3 over T1--T3,
complementing the simulation $d_{\max}$ metric.

Across all three hardware tasks, improvement emerges over ordinary repeated
task executions with fixed policy weights. The matched simulation comparisons
provide the controlled attribution to cross-trial context; the hardware results show that the corresponding repeated-improvement behavior persists through the deployed sensing--control loop.

\endgroup
\raggedbottom

\FloatBarrier

\section{Conclusion}
\label{sec:conclusion}
We presented RopeFormer, a history-conditioned framework that retains ordered
state--action--response history across trials with fixed policy weights. In matched simulation
comparisons, cross-trial context improved subsequent control across sustained
single-arm rotation, bimanual rotation, and transient whipping, although the
magnitude of the benefit varied with rope dynamics and observation settings.
On the Unitree H1-2, the same fixed-weight policies showed repeated
improvement on previously unseen physical ropes across all three tasks.
Together, these results show that ordinary task interaction can provide useful
context for adapting to unknown rope dynamics without explicit online
parameter estimation or policy updates. The current evaluation is limited to three rope-manipulation tasks and short multi-trial horizons; extending this principle to broader deformable objects, longer interaction histories, and richer sensing remains an important direction for future work.

\bibliographystyle{IEEEtran}
\bibliography{references}

\end{document}